\pdfoutput=1

\documentclass[11p]{article}

\usepackage[]{acl}

\usepackage{times}
\usepackage{latexsym}
\usepackage{amssymb}
\usepackage{hyperref}
\usepackage{booktabs}
\usepackage{subfig}
\usepackage{floatrow}
\usepackage{multirow}
\usepackage{graphicx}
\usepackage{amsmath}
\usepackage{soul}
\usepackage[normalem]{ulem}
\usepackage{colortbl}
\usepackage{wrapfig}
\usepackage{dialogue}
\usepackage{listings}
\usepackage{xcolor}
\usepackage[T1]{fontenc}

\usepackage[utf8]{inputenc}

\usepackage{microtype}

\title{Quantized Side Tuning: Fast and Memory-Efficient Tuning of\\Quantized Large Language Models}

\author{%
  Zhengxin Zhang$^{\ddagger\mathsection}$, Dan Zhao$^\flat$, Xupeng Miao$^\ddagger$,  Gabriele Oliaro$^\ddagger$ \\
   \textbf{Zhihao Zhang}$^\ddagger$\textbf{,} \textbf{Qing Li}$^\flat$\textbf{,} \textbf{Yong Jiang}$^\natural$\textbf{,} \textbf{Zhihao Jia}$^\ddagger$
   \\
   $^\ddagger$Carnegie Mellon University, $^\mathsection$Tsinghua University, \\$^\flat$Peng Cheng Laboratory, $^\natural$Tsinghua Shenzhen International Graduate School \\
  \texttt{zhang-zx21@mails.tsinghua.edu.cn, zhihao@cmu.edu}\\
}

\begin{document}
\maketitle
\begin{abstract}
Finetuning large language models (LLMs) has been empirically effective on a variety of downstream tasks. Existing approaches to finetuning an LLM either focus on parameter-efficient finetuning, which only updates a small number of trainable parameters, or attempt to reduce the memory footprint during the training phase of the finetuning. Typically, the memory footprint during finetuning stems from three contributors: model weights, optimizer states, and intermediate activations.
However, existing works still require considerable memory and none can simultaneously mitigate memory footprint for all three sources.
In this paper, we present Quantized Side Tuing (QST), which enables memory-efficient and fast finetuning of LLMs by operating through a dual-stage process. First, QST quantizes an LLM's model weights into 4-bit to reduce the memory footprint of the LLM's original weights; QST also introduces a side network separated from the LLM, which utilizes the hidden states of the LLM to make task-specific predictions. Using a separate side network avoids performing backpropagation through the LLM, thus reducing the memory requirement of the intermediate activations. Furthermore, QST leverages several low-rank adaptors and gradient-free downsample modules to significantly reduce the trainable parameters, so as to save the memory footprint of the optimizer states. 
Experiments show that QST can reduce the total memory footprint by up to 2.3 $\times$ and speed up the finetuning process by up to 3 $\times$ while achieving competent performance compared with the state-of-the-art. When it comes to full finetuning, QST can reduce the total memory footprint up to 7 $\times$.

\end{abstract}

\section{Introduction}
\begin{figure*}
    \centering
    \subfloat[Comparing the memory requirement of different methods.]{\label{fig:mem_comp_70b}
    \includegraphics[width=0.52\linewidth]{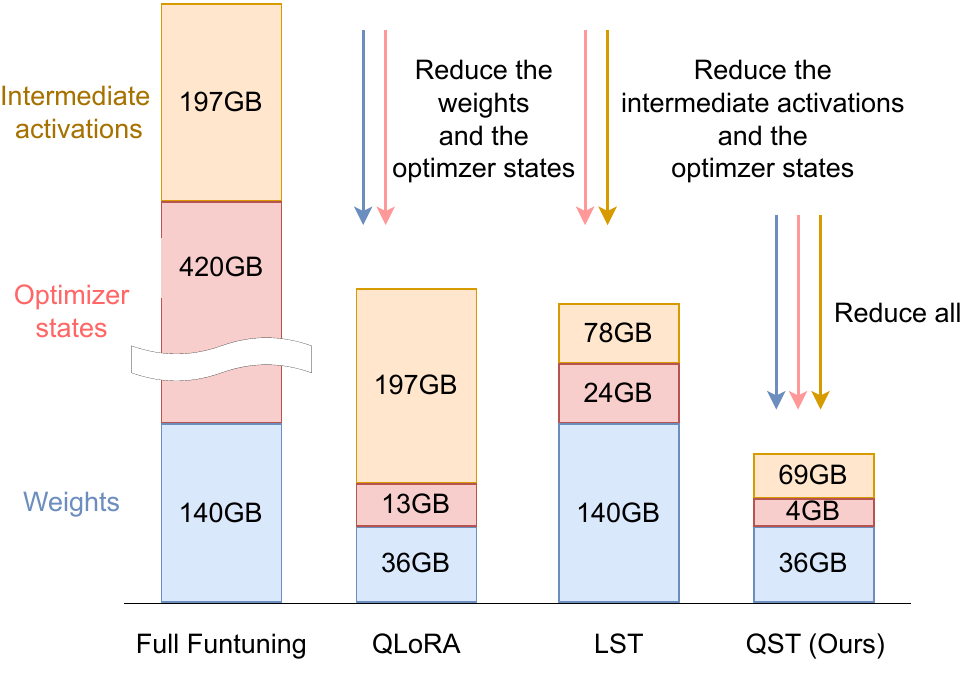}}
    \subfloat[Comparing the predictive performance of different finetuning techniques.]{\label{fig:acc_comp_7ob}
    \includegraphics[width=0.44\linewidth]{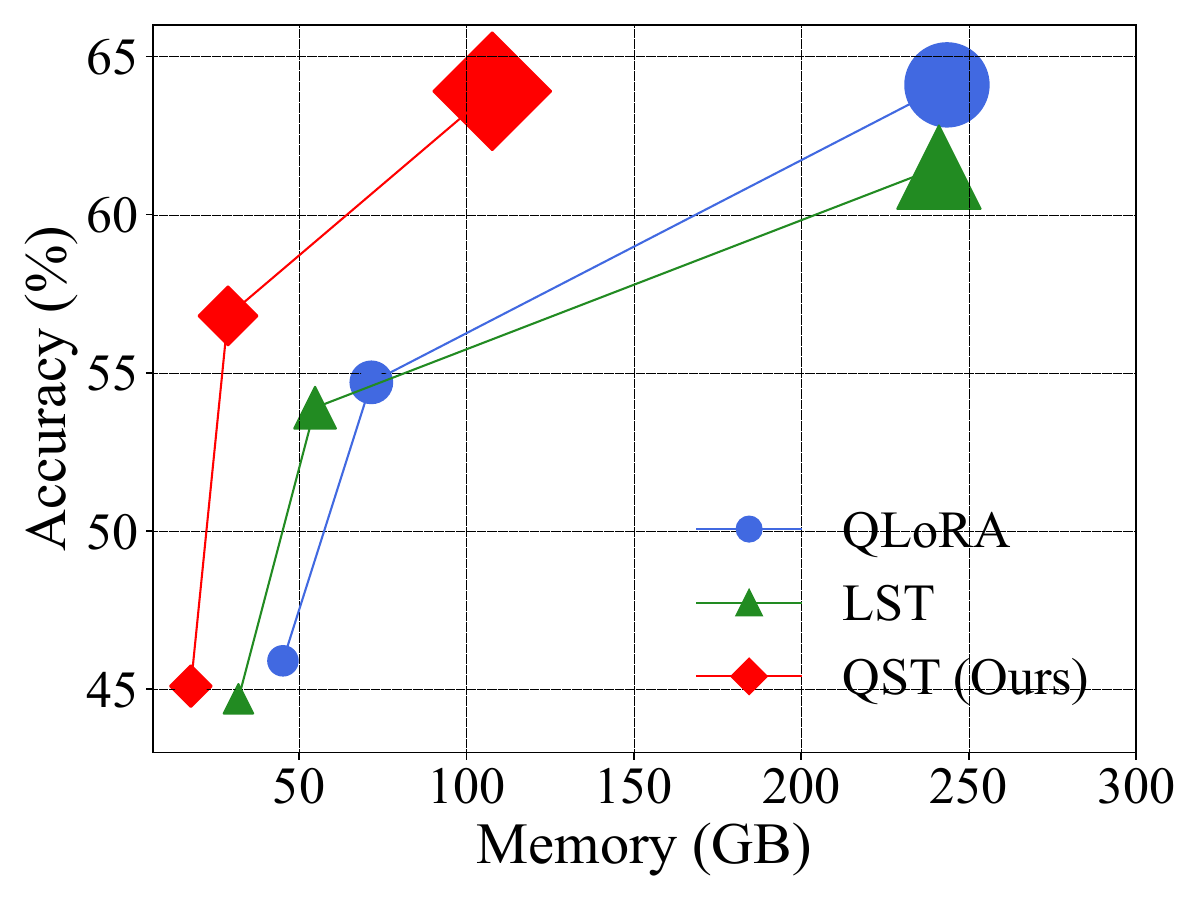}}
    \caption{Figure \ref{fig:mem_comp_70b} shows the memory footprint of different methods of fintuning LLaMA-2-70b. Figure \ref{fig:acc_comp_7ob} shows the MMLU 5-shot accuracy of different methods when tuning LLaMA-2-7B,  LLaMA-2-13B, and LLaMA-2-70B.
    Note that we set the batch size to 16 and the sequence length to 384. Larger markers represent larger models.
    }
    \label{fig:memory-70b}
\end{figure*}
Recent advancements in large language models (LLMs), including GPT~\citep{brown2020language,floridi2020gpt,openai2023gpt4}, PaLM \citep{chowdhery2022palm}, OPT \citep{zhang2022opt}, and LLaMA \citep{touvron2023llama}, have showcased remarkable task-generalization capabilities across diverse applications \citep{stiennon2020learning,dosovitskiy2020image}. 
The ongoing evolution of LLMs' capabilities is accompanied by exponential increases in LLMs' sizes, with some models encompassing 100 billion parameters \citep{raffel2020exploring,scao2022bloom}.
Finetuning pre-trained LLMs~\citep{min2021metaicl,wang2022super,wang2022self,liu2022few} for customized downstream tasks provides an effective approach to introducing desired behaviors, mitigating undesired ones, and thus boosting the LLMs' performance~\citep{ouyang2022training,askell2021general,bai2022training}.
Nevertheless, the process of LLM finetuning is characterized by its substantial memory demands. For instance, finetuning a 16-bit LLaMA model with 65 billion parameters requires more than 780GB of memory~\citep{dettmers2023qlora}. 

To reduce the computational requirement of LLM finetuning, recent work introduces {\em parameter-efficient finetuning} (PEFT), which updates a subset of trainable parameters from an LLM or introduces a small number of new parameters into the LLM while keeping the vast majority of the original LLM parameters frozen~\citep{houlsby2019parameter,li2021prefix,pfeiffer2020adapterfusion,hu2021lora,he2021towards,lester2021power}.
PEFT methods achieve comparable performance as full finetuning while enabling fast adaption to new tasks without suffering from catastrophic forgetting \citep{pfeiffer2020adapterfusion}. 
However, PEFT methods necessitate caching intermediate activations during forward processing, since these activations are needed to update trainable parameters during backward propagation. 
As a result, PEFT methods require saving more than 70\% of activations and almost the same training time compared to full finetuning~\citep{liao2023make,sung2022lst}.
Concisely, existing PEFT techniques cannot effectively reduce the memory footprint of LLM finetuning, restricting their applications in numerous real-world memory-constrained scenarios.

Recent work has also introduced approaches to combining PEFT and quantization. For example, QLoRA \citep{dettmers2023qlora} quantizes an LLM's weights to 4-bit and leverages low-rank adaption (LoRA)~\citep{he2021towards} to finetune the quantized LLM.
QLoRA reduces the memory footprint of an LLM's weights and optimizer states, and as a result, finetuning a 65B LLM requires less than 48 GB of memory.
However, QLoRA does not consider the memory footprint of intermediate activations, which can be particularly large when using a large batch size for finetuning.
As a result, QLoRA only supports small-batch training (e.g. a batch size of $1$), and finetuning a 65B LLM requires checkpointing gradients~\cite{chen2016training} to fit the LLM on a single 48GB GPU, resulting in long training time.
Besides, our evaluation also reveals that the performance of QLoRA becomes unstable when using 16-bit floating points.
\citet{sung2022lst} and \citet{zhang2020side} propose to use a {\em side network} to reduce the memory footprint of intermediate activations by {\em avoiding backpropagation of the LLM} on natural language processing (NLP) and computer vision (CV) tasks, respectively.
Even with the adoption of a side network, the inherent model size of the LLM remains a challenge. Meanwhile, these approaches focus on small models (i.e., less than 3 billion parameters), and their applicability and efficacy for larger models remain unexplored. 


\begin{figure*}
    \centering
    \includegraphics[width=\linewidth]{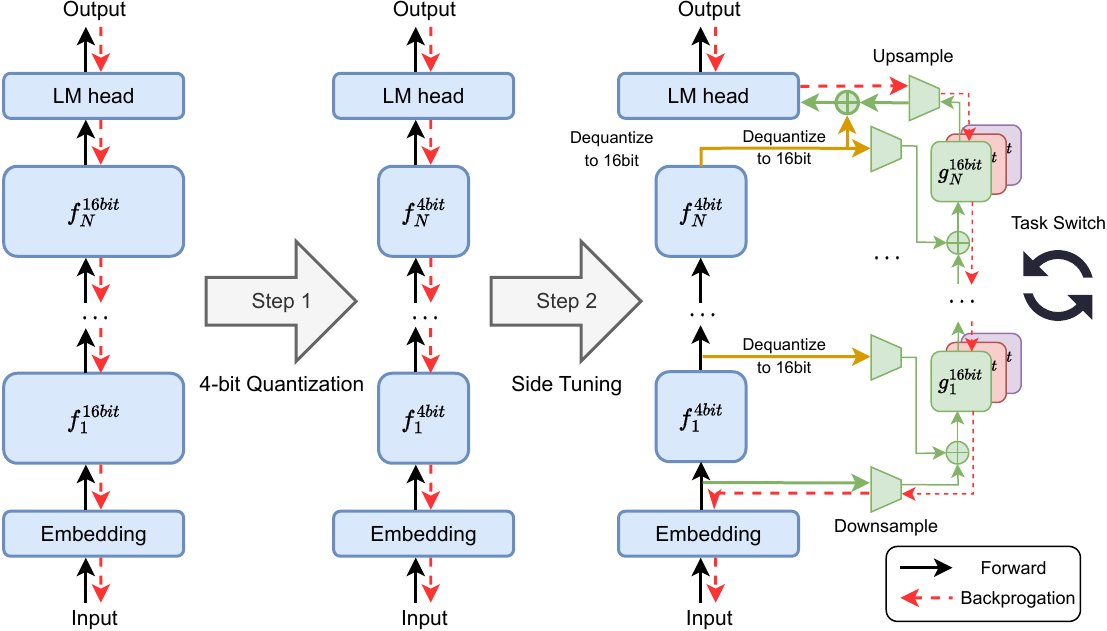}
    \caption{A overview of quantized side tuning.}
    \label{fig:qst}
\end{figure*}

In this paper, we propose a fast, memory-efficient LLM finetuning framework, called \textbf{Q}uantized \textbf{S}ide-\textbf{T}uning (QST), which operates through a dual-stage process as shown in Figure~\ref{fig:qst}.
First, QST quantizes an LLM into 4-bit to reduce the memory footprint of its model weights. 
Second, QST introduces a side network separating from
the quantized LLM to avoid performing backward propagation for the quantized LLM, thus saving the memory footprint of intermediate activations.
During the training phase of QST, the input to each layer of the side network is formed by combining (1) the downsampled output of the corresponding quantized LLM layer and (2) the output of the previous layer of the side network.
A larger LLM usually has a larger model depth (i.e., the number of layers) and width (the hidden size of each layer), which in turn requires more trainable parameters for the downsampling layers. 
Unlike \citet{sung2022lst} that leverages linear layer to perform downsampling, QST uses several low-rank adapter methods \citep{he2021towards,edalati2022krona} such as MaxPooling \citep{lecun1998gradient} and AvgPooling, significantly reducing the required trainable parameters and the memory footprint for the optimizer states.
After that, we use a learnable parameter to assign weights and subsequently aggregate the hidden states of the quantized LLM and the side network. 
Finally, we reuse the LLM head or classifier to predict.
Combined with 4-bit quantization and side tuning, QST significantly reduces all three main contributors of the memory footprint and training time during the training phase. Besides, QST does not increase inference latency since the LLM and side network can be computed in parallel. Figure \ref{fig:memory-70b} compares the memory footprint of QST and existing parameter-efficient fine-tuning methods, including QLoRA and LST.

To validate the effectiveness of our QST, we conduct extensive evaluations for different types of LLMs (e.g., OPT, LLaMA 2), with 1.3B to 70B parameters, on various benchmarks. Experiment results show that QST can reduce the total memory footprint by up to 2.3 $\times$ and speed up the finetuning process by up to 3 $\times$ while achieving competent performance compared with the state-of-the-art. 
Our codes are released to the GitHub \footnote{\href{https://github.com/YouAreSpecialToMe/QST}{https://github.com/YouAreSpecialToMe/QST}}
.

\section{Related Work}
\subsection{Parameter-Efficient Finetuning}
Finetuning allows an LLM to adapt to specialized domains and tasks \citep{devlin2018bert,radford2019language,brown2020language}. However, fully finetuning an LLM comes with high computation costs due to the rapidly increasing LLM sizes. {\em Parameter-efficient finetuning} (PEFT) methods are proposed to solve this issue. Drawing inspiration from the pronounced sensitivity of LLMs to prompts as highlighted in \citet{schick2020exploiting}, a series of studies introduce trainable prompt embeddings prepended to the input text or attention components while preserving the original LLM parameters~\cite{liu2023gpt,li2021prefix,lester2021power}. 
\citet{rusu2016progressive} and \citet{houlsby2019parameter} propose {\em adapter} modules to introduce new task-specific parameters, which are inserted into the Transformer layers inside the LLM. LoRA~\cite{hu2021lora} leverages the low-rank decomposition concept to construct trainable parameters inserted into the original LLM weights. (IA)$^3$ \cite{liu2022few} proposes to scale the pre-trained model weights of an LLM with a trainable vector. Of late, there has been a surge in the proposal of unified approaches that amalgamate various PEFT methods by leveraging human heuristics \cite{he2021towards} or employing neural architecture search \cite{zhou2023autopeft,zoph2016neural,mao2021unipelt}. Existing PEFT approaches focus on optimizing model performance while minimizing trainable parameters. However, a reduction in the number of trainable parameters does not inherently imply a corresponding reduction in memory footprint.

\subsection{Memory-Efficient Training and Finetuning}
Memory-efficient training and finetuning aims to reduce the memory footprint during the LLM training and/or finetuning phase. Reversible neural networks \cite{gomez2017reversible,kitaev2020reformer,mangalam2022reversible} allow the intermediate activations of each layer to be recomputed from the activation of its next layer, thus exempting the need to save intermediate activations. {\em Gradient checkpointing}~\cite{chen2016training} offers an optimization strategy that balances computational resources against memory footprint. Specifically, it reduces memory requirement by selectively discarding certain intermediate activations, which are subsequently recomputed through an additional forward pass when needed. Another line to enhancing memory efficiency involves network compression, that is, the original LLM is reduced to a more compact form, thereby making both the training and inference phases more computationally economical. Network pruning and distillation are the most prevalent strategies for network compression. Network distillation \cite{hinton2015distilling,koratana2019lit} involves the creation of a student network that is trained to approximate the output distribution of a teacher network across a specified dataset. Network pruning \cite{frankle2018lottery,frankle2020linear} aims to streamline models by ascertaining the significance of individual parameters and subsequently eliminating those deemed non-essential. Compared with PEFT methods, network compression yields models optimized for expedited inference, whereas PEFT methods may achieve superior performance by updating a small set of trainable parameters.

Recently, QLoRA \cite{dettmers2023qlora} quantizes the LLM to 4-bit and then adds LoRA to finetune the quantized LLM. QLoRA significantly reduces the memory footprint of weights and optimizer states compared with full finetuning while retaining similar performance.  QLoRA does not consider the memory footprint of intermediate activations, and thus falls short in finetuning the LLM with a large batch size, resulting in a long training time.
In the context of NLP and CV tasks, the studies by \cite{sung2022lst} and \cite{zhang2020side} introduce the concept of employing a side network. The side network aims to obviate the need for backpropagation through the LLM, thereby reducing the memory footprint associated with intermediate activations. Despite incorporating the side network, the inherent model size (i.e., the memory footprint of weights) of the LLM still poses computational challenges. Hence, both methods can only focus exclusively on models with fewer than 3 billion parameters, and fail to finetune models with more parameters. 
\section{Quantized Side Tuning}
In this section, we first describe the process of quantizing an LLM into 4-bit, and then introduce our design of the side network for side tuning.
\subsection{4-bit Quantization}
Quantization is the process of converting a data type with more bits (e.g., 32- or 16-bit floating points) into another data type with fewer bits (e.g., 8-bit integers or 4-bit floating points). QST first quantizes an LLM from 16-bit into 4-bit, formulated as follows.
\begin{align}
    X^{4bit} = & \text{round}\left(\frac{M_{4bit}}{\text{Absmax}(X^{16bit})}X^{16bit}\right) \\
    = & \text{round}\left(c^{16bit} \cdot X^{16bit}\right),
\end{align}
where $X^{4bit}$ and $X^{16bit}$ are tensors in 4- and 16-bit, respectively. $M_{4bit}$ is the maximum value of the 4-bit data type. For example, $M_{NF4}=1$, where NF4 is an information-theoretically optimal data type that ensures each quantization bin has an equal number of values assigned from the input tensor. QST considers both NF4 and FP4 to quantize an LLM. We empirically demonstrate that NF4 performs the best in our experiments (see Section \ref{exp:4-bit}).
$c^{16bit}$ is the quantization constant (or quantization scale) of the 16-bit data type. Correspondingly, dequantization is given by 
\begin{align}
    dequant(c^{16bit},X^{4bit}) = \frac{X^{4bit}}{c^{16bit}} = X^{16bit}.
\end{align}
The key limitation of this method arises when the input tensor contains values with very large magnitudes, commonly referred to as {\em outliers}.
Such outliers can result in under-utilization of the quantization bins, leading to sparsely populated or even empty bins in some instances. To address this issue, a prevalent strategy involves partitioning the input tensor into discrete blocks, each subjected to independent quantization with its own associated quantization constant. As a result, the input tensor $ X \in \mathbb{R}^{b \times h}$ is decomposed into $n$ contiguous blocks, each comprising $B$ elements. This decomposition is facilitated by flattening $X$ into a 1-dimensional array, which is then partitioned into $n = \frac{(b \times h)}{B}$ individual blocks. Then, we can leverage E.q. (1) to independently quantize these $n$ blocks using different quantization constants. Typically, minimizing the error associated with 4-bit quantization would necessitate the utilization of smaller block sizes. This is attributed to the reduced influence of outliers on other weights. However, using a small block size leads to high memory overhead since we need to allocate more memory for these quantization constants. To reduce the memory footprint of quantization constants, we can use the same quantization strategy to quantize these quantization constants~\citep{dettmers2023qlora}. In this paper, we use 8-bit float points to quantize the quantization constants, and the forward pass of a single linear layer in the LLM is defined as $Y^{16bit} = dequant(dequant(c^{16bit}_2,c^{8bit}_1), W^{4bit})X^{16bit}$.
4-bit quantization can significantly reduce the memory footprint of weights, facilitating easier storage and deployment of LLMs. Besides, low-precision floating numbers are faster to execute on modern accelerators such as GPUs, leading to faster model training and inference. 
Nonetheless, the high to low precision data type conversion process during quantization can lead to accuracy degradation, attributable to the inherent information loss. 

\subsection{Side Tuning}
We now analyze the memory footprint of LLM training and then introduce the neural architecture of the side network, which reduces the inherent information loss and minimizes accuracy drop during quantization.

\paragraph{Memory footprint during the training phase.} 
For a given LLM with $N$ layers, let $y_i = f_i(W_i, x_i)$ denotes the $i^{th}$ transformer layer of the LLM, where $x_i$ is the input to the $i^{th}$ layer (i.e., $x_i=y_{i-1}$).  
The memory required during the training phase of the LLM predominantly comprises three main contributors: \textbf{M1}) weights of the LLM $\{W_i\}_{i=1}^{N}$, \textbf{M2}) the optimizer state, which is threefold the size of the trainable parameters when employing the Adam optimizer \citep{kingma2014adam} (one for gradient and two for moments), and \textbf{M3}) the intermediate activations $\{y_{i}^\prime\}_{i=1}^N$. The memory footprint of intermediate activations is related to model depth, width, and several training settings, e.g., batch size and sequence length.
QLoRA reduces the memory footprint of an LLM's weights and optimizer states (M1 and M2) but fails to reduce intermediate activations (M3). 
When finetuning an LLM with a large batch size and/or long sequence length, the memory footprint of QLoRA increases significantly. However, using a small batch size results in long training time. \citet{sung2022lst} only reduces the memory footprint of intermediate activations (M3), thus it struggles to finetune a model with more than 3 billion parameters. 

\begin{figure}
    \centering
    \includegraphics[width=\linewidth]{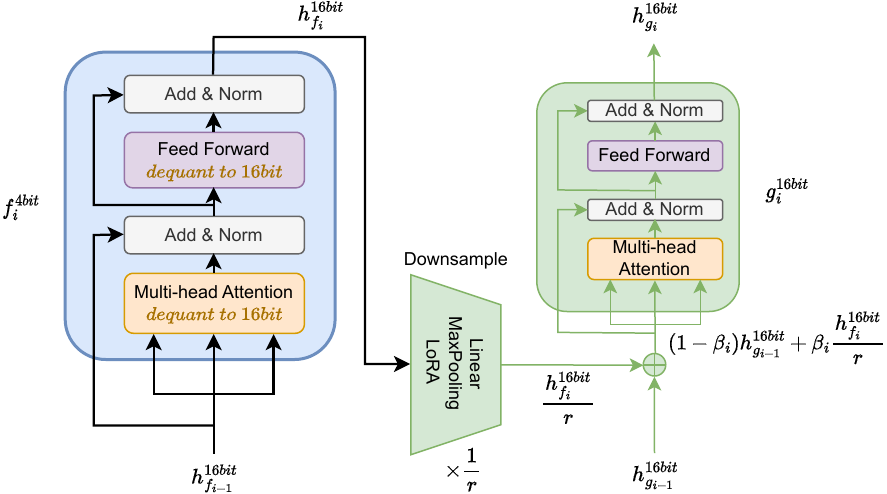}
    \caption{Illustration of $i^{th}$ layer of QST.}
    \label{fig:qst_layer}
\end{figure}
\paragraph{Side network.} 
Our side network $g$ serves as a lightweight version of the quantized LLM $f$. The hidden state and weight dimension of $g$ are $r$ times smaller than those of $f$, where $r$ is the reduction factor. 
During the forward pass, the hidden state of the $i^{th}$ layer of the side network 
$h_{g_i}$ is formulated by $h^{16bit}_{g_i} = (1-\beta_i)*downsample_i(h^{16bit}_{f_i}) + \beta_i*h^{16bit}_{g_{i-1}},$
where $h^{16bit}_{f_i}$ is the hidden state of the $i^{th}$ layer of $f$ and can be computed using E.q. (3). The illustration of $i^{th}$ layer of our QST is shown in Figure \ref{fig:qst_layer}. Note that we use the output of the embedding layer and the downsampled embedding layer as $h^{16bit}_{f_0}$ and $h^{16bit}_{g_0}$. $\beta_i = sigmoid(\gamma_i)$ is a learned gate parameter of $i^{th}$ layer, where $\gamma_i$ is a learnable zero-initialized scalar. $downsample_i$ is the downsample module of the $i^{th}$ layer to reduce the hidden state dimension of $f$ by $r$ times. Prior work leverages linear projections to downsample (i.e., $\times \frac{1}{r}$) the high-dimensional hidden states of $f$ to the low-dimensional hidden states of $g$. 
However, an LLM typically comprises plenty of layers with substantially high-dimensional hidden states, particularly when the number of parameters exceeds 3 billion. Using linear projections to downsample involves a significant amount of trainable parameters, requiring a high memory footprint for the parameters and their optimizer states. For example, if the LLM has 24 layers, the dimension of its hidden state is 2048 and the reduction factor $r$ is 4, the downsample module consumes about 50\% of the overall trainable parameters. 

To address this problem, we leverage several different downsample methods, including LoRA \cite{he2021towards}, Adapter \cite{edalati2022krona}, MaxPooling \cite{lecun1998gradient} and AvgPooling. 
LoRA augments a linear projection through an additional factorized projection, which can be formulated as $W=L_1L_2$, where $W \in \mathbb{R}^{d_{in}\times d_{out}}$, $L_1 \in  \mathbb{R}^{d_{in} \times d_{r}}$ and $L_2 \in  \mathbb{R}^{d_{r} \times d_{out}}$.
Adapter is similar to LoRA but introduces an extra non-linear function between $L_1$ and $L_2$.
Using LoRA or Adapter can reduce the ratio of the trainable parameters of these downsample modules from 56\% to 8\%. 
MaxPooling and AvgPooling do not introduce extra trainable parameters. We empirically demonstrate that the Adapter performs the best in our experiments. 

Finally, we upsample (i.e., $\times r$) from low-dimensional hidden states of $g$ to high-dimensional hidden states of $f$. 
Prior works have only evaluated side tuning methods (e.g., LST) in the context of classification tasks. We observe that LST suffers from repetition when generating long-sequence texts, which renders it incapable of producing extensive and high-quality texts. This limitation stems from LST's utilization of the hidden states of the side network for prediction, which causes an initialization position far removed from the pre-trained model at the onset of finetuning. 
As illustrated in MEFT \citep{liao2023make}, the initialization step emerges as a critical factor influencing the efficacy of fine-tuning methods.
To resolve this issue, QST combines the output of the LLM's last layer (i.e., $h^{16bit}_{f_N}$) with the output of the side network (i.e., $h^{16bit}_{g_N}$), and sends the weighted sum $h^{16bit}_N = \alpha h^{16bit}_{f_N} + (1 -\alpha)h^{16bit}_{g_N}$ to the LM head, where $\alpha$ is a learnable parameter. We initialize $\alpha$ to 1 to preserve the starting point from the pre-trained model at the beginning of finetuning, which is consistent with the initialization of LoRA.
With this design, when switching across different downstream tasks, QST can fulfil the necessary adjustments by altering the side network alone, and thus obviating the need for redeploying the LLM.

QST only updates the parameters of the side network $g$, but not the 4-bit weights in the LLM $f$. Unlike QLoRA, the calculation of the gradient $\frac{\partial L}{\partial g}$ does not entail the calculation of $\frac{\partial L}{\partial f}$, thus avoiding the extensive computational costs of performing backpropagation on $f$, which ultimately reduces the memory footprint of intermediate activations and speeds up finetuning.

In summary, QST leverages a 4-bit data type to store an LLM's model weights, thus reducing the memory footprint of weights (M1). In addition, QST leverages a 16-bit computation data type for the forward pass and backpropagation computation and only computes the gradient of weights in $g$ (M3). Finally, QST leverages several factorized projection and gradient-free downsample methods to reduce the trainable parameters (M2). These techniques together allow QST to reduce the memory requirement for all three factors, resulting in fast and memory-efficient finetuning with a nearly 1\% performance drop.

\section{Evaluation}
In this section, we empirically validate the effectiveness of our QST method by examining its performance for LLMs with different types (e.g., OPT and LLaMA 2), sizes (from 1.3B to 70B), and benchmarks.

\begin{table*}[h]
\caption{Experiments results on GLUE benchmark (using BF16 data type).}
\label{tab:res_glue}
\resizebox{\textwidth}{!}{
\begin{tabular}{@{}cccccccccccc@{}}
\toprule
Method  & \# Param. (\%) & Memory (GB) & RTE & MRPC & STS-B & CoLA & SST-2 & QNLI & QQP & MNLI & Avg. \\ \midrule
\multicolumn{12}{c}{\cellcolor[HTML]{C0C0C0}OPT-1.3B (batchsize=16, sequence length=512)}                              \\
QLoRA   &        4.41\%         &     31.3       &  81.3$_{\pm 1.6}$   &   \ul{83.3}$_{\pm 1.1}$   &   \textbf{89.9}$_{\pm 0.5}$    &   \ul{62.1}$_{\pm 2.3}$  &  \textbf{94.9}$_{\pm 0.1}$     &  \textbf{86.3}$_{\pm 0.2}$     &  \textbf{87.1}$_{\pm 0.1}$  &  76.0$_{\pm 0.3}$     &  \textbf{82.6}    \\
LST     &       2.39\%         &    \ul{20.9}       &  82.0$_{\pm 2.2}$   &   83.1$_{\pm 1.3}$   &   88.6$_{\pm 0.4}$    &   59.5$_{\pm 3.1}$   &   \ul{94.4}$_{\pm 0.3}$    &   \ul{86.1}$_{\pm 0.3}$    &   86.4$_{\pm 0.6}$  &   77.8$_{\pm 0.5}$   &   82.2   \\
LoRA    &       2.36\%         &     32.9      &  \textbf{82.7}$_{\pm 1.9}$   &   \textbf{83.4}$_{\pm 0.9}$   &    89.3$_{\pm 0.2}$   &   \textbf{62.5}$_{\pm 1.7}$    &   93.7$_{\pm 0.7}$   &   81.4$_{\pm 9.3}$    &  \ul{86.9}$_{\pm 0.3}$   &   \textbf{81.2}$_{\pm 0.1}$   &  \textbf{82.6}    \\
Adapter &        \ul{0.48\%}        &      32.5     &  \ul{82.2}$_{\pm 0.8}$   &   82.7$_{\pm 1.4}$   &   \ul{89.7}$_{\pm 1.6}$    &   60.6$_{\pm 3.0}$   &   93.8$_{\pm 0.2}$    &   83.6$_{\pm 0.1}$     &   86.3$_{\pm 0.4}$  &   \ul{80.5}$_{\pm 0.1}$   &   \ul{82.4}   \\ 
QST    &       \textbf{0.45\%}         &       \textbf{17.7}      &  79.5$_{\pm 2.5}$   &   81.7$_{\pm 1.1}$   &   88.4$_{\pm 1.1}$    &    59.7$_{\pm 2.9}$  &   94.3$_{\pm 0.3}$    &  85.7$_{\pm 0.5}$     &  84.3$_{\pm 0.7}$   &  77.1$_{\pm 0.6}$    &   81.3   \\\bottomrule
\toprule
\multicolumn{12}{c}{\cellcolor[HTML]{C0C0C0}OPT-2.7B (batchsize=16, sequence length=512)}                              \\
QLoRA   &     3.57\%           &      47.0          &  83.6$_{\pm 1.5}$   &  \textbf{84.8}$_{\pm 1.2}$    &   \ul{91.2}$_{\pm 0.6}$    &    \ul{63.7}$_{\pm 2.6}$  &   \textbf{95.6}$_{\pm 0.2}$    &  \textbf{88.7}$_{\pm 0.1}$     &  89.5$_{\pm 0.2}$   &  78.3$_{\pm 0.4}$    &    \ul{84.4}  \\
LST     &     2.39\%           &      \ul{30.7}         &  82.5$_{\pm 2.9}$   &   83.9$_{\pm 1.5}$   &   89.1$_{\pm 0.9}$    &   60.7$_{\pm 3.5}$   &    95.3$_{\pm 0.4}$   &   \ul{87.3}$_{\pm 0.2}$    &  88.8$_{\pm 1.0}$   &  80.4$_{\pm 0.7}$    &   83.5   \\
LoRA    &      1.90\%          &    50.4        &    \textbf{84.7}$_{\pm 1.4}$  &   \ul{84.6}$_{\pm 0.8}$   &   90.9$_{\pm 0.1}$    &   \textbf{64.5}$_{\pm 2.4}$   &   95.3$_{\pm 0.6}$    &   83.0$_{\pm 7.4}$    &  \textbf{90.7}$_{\pm 0.1}$   &   \textbf{82.6}$_{\pm 0.2}$   &   \textbf{84.5}   \\
Adapter &       \textbf{0.37\%}         &     49.9   &  \ul{84.4}$_{\pm 0.7}$   &   83.7$_{\pm 1.4}$   &   \textbf{91.5}$_{\pm 1.9}$    &  63.4$_{\pm 3.8}$    &   \ul{95.4}$_{\pm 0.3}$    &  83.6$_{\pm 0.2}$     &  \ul{90.2}$_{\pm 0.3}$   &   \ul{81.1}$_{\pm 0.1}$   &   84.2   \\ 
QST    &      \ul{0.43\%}          &      \textbf{24.4}             &  80.1$_{\pm 2.1}$   &   83.7$_{\pm 1.2}$   &   88.9$_{\pm 1.4}$    &    62.0$_{\pm 3.4}$  &   95.2$_{\pm 0.8}$    &  86.6 $_{\pm 0.9}$    &  86.5$_{\pm 0.9}$   &  80.4$_{\pm 0.6}$    &   83.0   \\\bottomrule
\toprule
\multicolumn{12}{c}{\cellcolor[HTML]{C0C0C0}OPT-6.7B (batchsize=16, sequence length=512)}                              \\
QLoRA   &        2.33\%         &    63.6  &  84.5$_{\pm 1.2}$   &   85.9$_{\pm 0.7}$   &   92.0$_{\pm 0.8}$    &   64.3$_{\pm 2.8}$    &   96.2$_{\pm 0.1}$    &   90.2$_{\pm 0.2}$    &   90.7$_{\pm 0.2}$  &   79.8$_{\pm 0.3}$   &    85.5  \\
QST    &     0.42\%           &      27.5               &  80.8$_{\pm 1.4}$   &  85.2$_{\pm 1.0}$    &    89.6$_{\pm 0.7}$   &   62.8$_{\pm 2.6}$   &   96.4$_{\pm 0.6}$    &  87.3$_{\pm 1.1}$     &  89.4$_{\pm 0.8}$   &   81.6$_{\pm 0.5}$   &   84.2   \\\bottomrule
\end{tabular}
}
\end{table*}

\subsection{Experimental Setup}
\textbf{Datasets.} We evaluate the performance of QST and several baselines on natural language understanding (NLU) and natural language generation tasks. For NLU experiments, we use the GLUE \cite{wang2018glue} (General Language Understanding Evaluation) and MMLU \cite{hendrycks2020measuring} (Massively Multitask Language Understanding) benchmarks. The GLUE benchmark provides a comprehensive evaluation of models across a range of linguistic tasks. These tasks encompass linguistic acceptability as examined in CoLA \cite{warstadt2019neural}, sentiment analysis as portrayed in SST2 \cite{socher2013recursive}, tasks probing similarity and paraphrase distinctions such as MRPC \cite{dolan2005automatically}, QQP \cite{Shankar1017quora}, and STS-B \cite{cer2017semeval}, in addition to natural language inference tasks including MNLI \cite{williams2017broad}, QNLI \cite{rajpurkar2016squad}, and RTE \cite{bentivogli2009fifth}. We report accuracy on MNLI, QQP, QNLI, SST-2, MRPC, and RTE, Pearson correlation coefficients on SST-B, and Mathews correlation coefficients \cite{matthews1975comparison} on CoLA. The MMLU benchmark consists of 57 tasks including elementary mathematics, US history, computer science, law, and more. We report the average 5-shot test accuracy on the 57 tasks.  

\noindent\textbf{Models.} We use decoder-only LLMs such as the OPT series (OPT-1.3B, OPT-2.7B, OPT-6.7B, OPT-13B, OPT-30B, and OPT-66B) and the LLaMA-2 series (LLaMA-2-7B, LLaMA-2-13B, and LLaMA-2-70B).

\noindent\textbf{Baselines.} We compare QST with QLoRA \cite{dettmers2023qlora}, LST \cite{sung2022lst}, LoRA \cite{he2021towards}, and Adapter \cite{houlsby2019parameter}. Note that we only compare LST, LoRA, and Adapter when the model size is less than 3B since their memory footprint of weights can be excessively huge beyond that. 

\noindent\textbf{Implementation.} We set the reduction factor $r$ to 16 by default.
We use Adapter as the downsample module, a linear layer as the upsample module, and set the rank of the Adapter to 16. 
We use the NF4 data type to store the weights of the LLM and bfloat16 as the data type for computation. 
We adopt the same parameters reported in QLoRA, LST, LoRA, and Adapter to construct the baselines. 
Other hyperparameters are specified in Appendix~\ref{sec:paramglue} and Appendix~\ref{sec:parammmlu}. We run each experiment three times under different random seeds and report the average performance.
We conduct all the experiments using Pytorch \cite{paszke2017automatic} and HuggingFace library \cite{wolf2019huggingface} on 4 NVIDIA RTX A5000 GPUs, each with 24GB memory.

\subsection{Experiments on GLUE Benchmark}
Table~\ref{tab:res_glue} shows the performance of different methods on the GLUE benchmark.
Overall, QST achieves the lowest memory footprint among all methods while attaining competent accuracy. 
Particularly, for relatively small models (i.e., OPT-1.3B and OPT-2.7B), QST reduces the memory footprint by around 2$\times$ compared with QLoRA, LoRA, and Adapter, while achieving comparable accuracy. Compared with LST, QST reduces the memory requirement by 3.2GB and 6.3GB for finetuning OPT-1.3B and OPT-2.7B. QST also reduces the trainable parameters by around 10$\times$ and 5$\times$ compared with QLoRA and the other baselines, respectively.

For larger models such as OPT-6.7B, we focus on comparing QST with QLoRA. This is because QLoRA has similar accuracy with the other baselines, but LoRA, Adapter, and LST all have excessively huge memory footprints of weights when it comes to finetuning OPT-6.7B\footnote{QLoRA can leverage gradient accumulation to finetune with a batch size of 16 while guaranteeing an affordable memory footprint.}.
Compared with QLoRA, QST reduces the memory footprint and trainable parameters by 2.3$\times$ and 5.5$\times$, while only introducing a 1.3\% accuracy drop.


\subsection{Experiments on MMLU Benchmark}
\begin{table*}[]
\caption{Experiment results (accuracy/memory) on MMLU 5-shot.}
\resizebox{\textwidth}{!}{
\begin{tabular}{@{}ccccccccccc@{}}
\toprule
Method & OPT-1.3B & OPT-2.7B & OPT-6.7B & OPT-13B & OPT-30B & OPT-66B & LLaMA-2-7B & LLaMA-2-13B & LLaMA-2-70B & Avg. \\ \midrule
QLoRA &      25.0/6.3    &   25.2/10.1       & 25.6/15.5     & 26.5/25.4   &   27.7/46.8      &   36.4/87.5      & 45.9/15.6  & 54.7/25.4   & 64.1/95.5  & 36.8/36.5\\
QST   &    24.3/3.2      &    25.5/4.8      & 26.2/7.2     &  26.8/12.6    &    27.3/25.7     &    36.0/52.3     & 45.1/7.3   & 56.8/12.6   & 63.9/56.0 &  36.9/20.2\\ \bottomrule
\end{tabular}
}
\end{table*}

The experiment results of the MMLU benchmark are shown in Table 2. We set the batch size to 4 and the sequence length to 384. We use the Alpaca dataset \citep{alpaca} to finetune both QLoRA and QST. We compare QST with QLoRA on accuracy and memory requirement over OPT-1.3B, OPT-2.7B, OPT-6.7B, OPT-13B, OPT-30B, OPT-66B, LLaMA-2-7B, LLaMA-2-13B, and LLaMA-2-70B. QST improves the accuracy by 0.1\% on average while reducing the memory footprint by 1.8$\times$ compared with QLoRA. Particularly, QST yields an enhancement of 2.1\% in accuracy over QLoRa when finetuning LLaMA-2-13B. When finetuning the OPT-2.7B, OPT-6.7B, and OPT-13B models, QST achieves 0.3\%, 0.6\%, and 0.3\% accuracy improvements, respectively. 

\subsection{Memory Footprint Analysis}
\label{sec:mfea}
\begin{figure*}[htbp]
\centering
\subfloat[Effect of batch size.]{\label{bs_compare}
\includegraphics[width=0.33\linewidth]{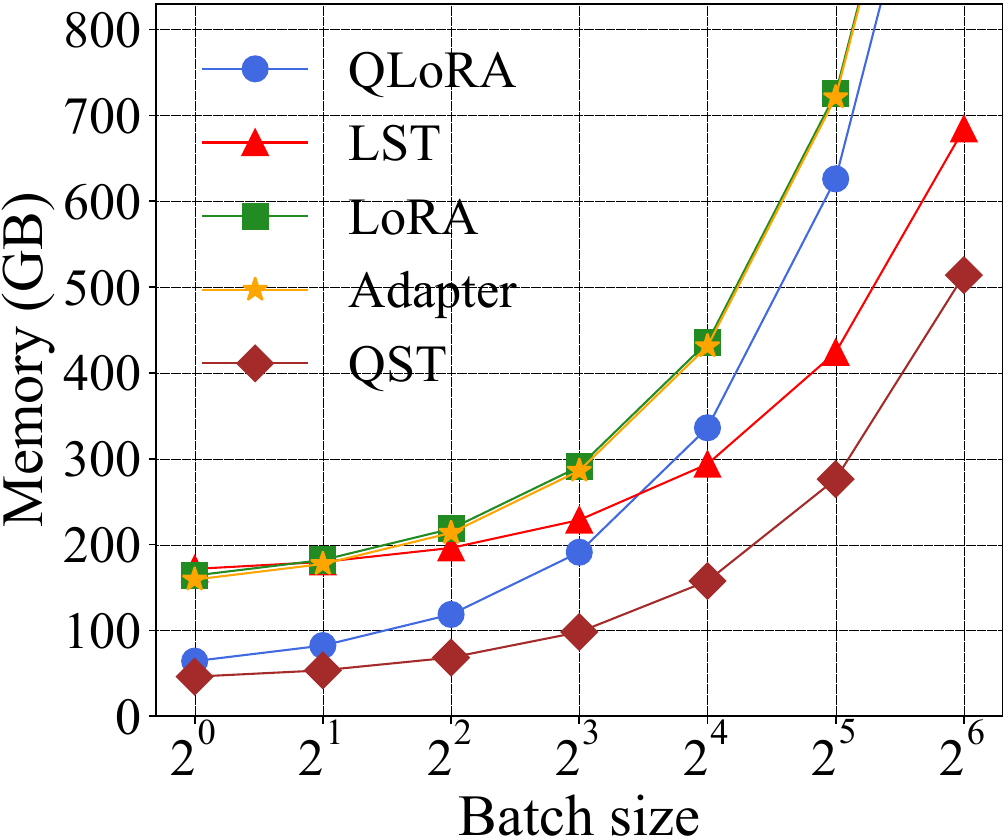}
}
\subfloat[Effect of total model bits.]{\label{ms_compare}
\includegraphics[width=0.33\linewidth]{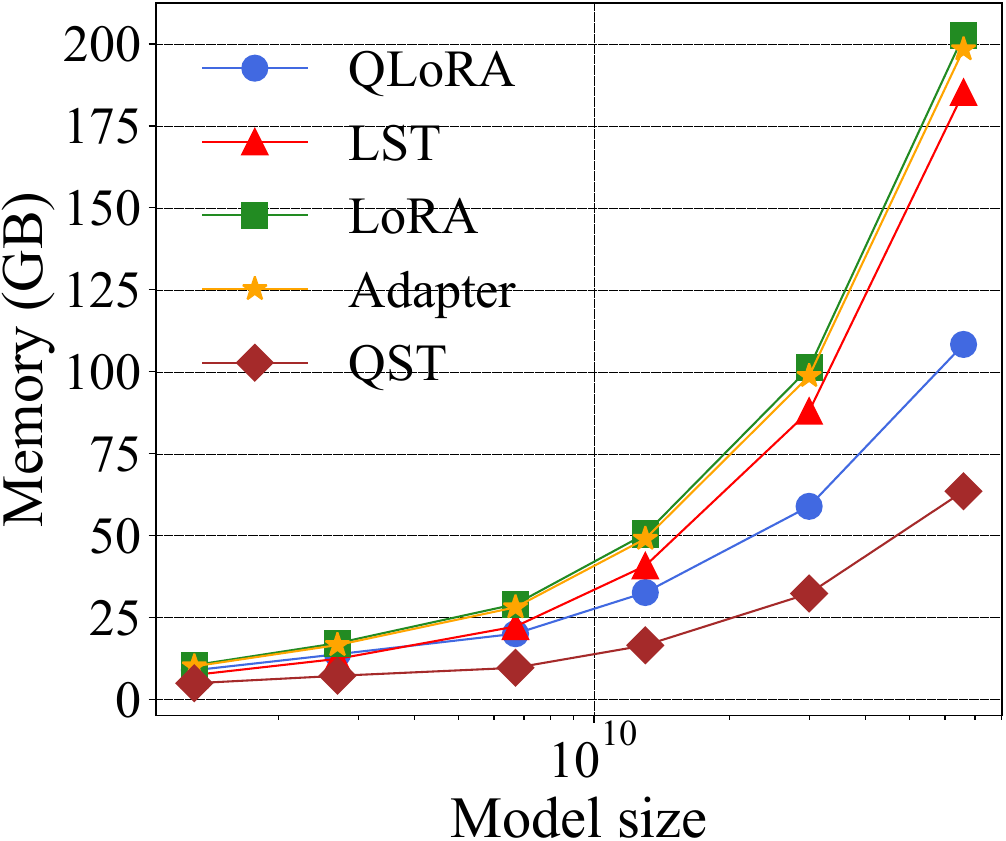}}
\subfloat[Effect of sequence length.]{\label{seq_compare}
\includegraphics[width=0.33\linewidth]{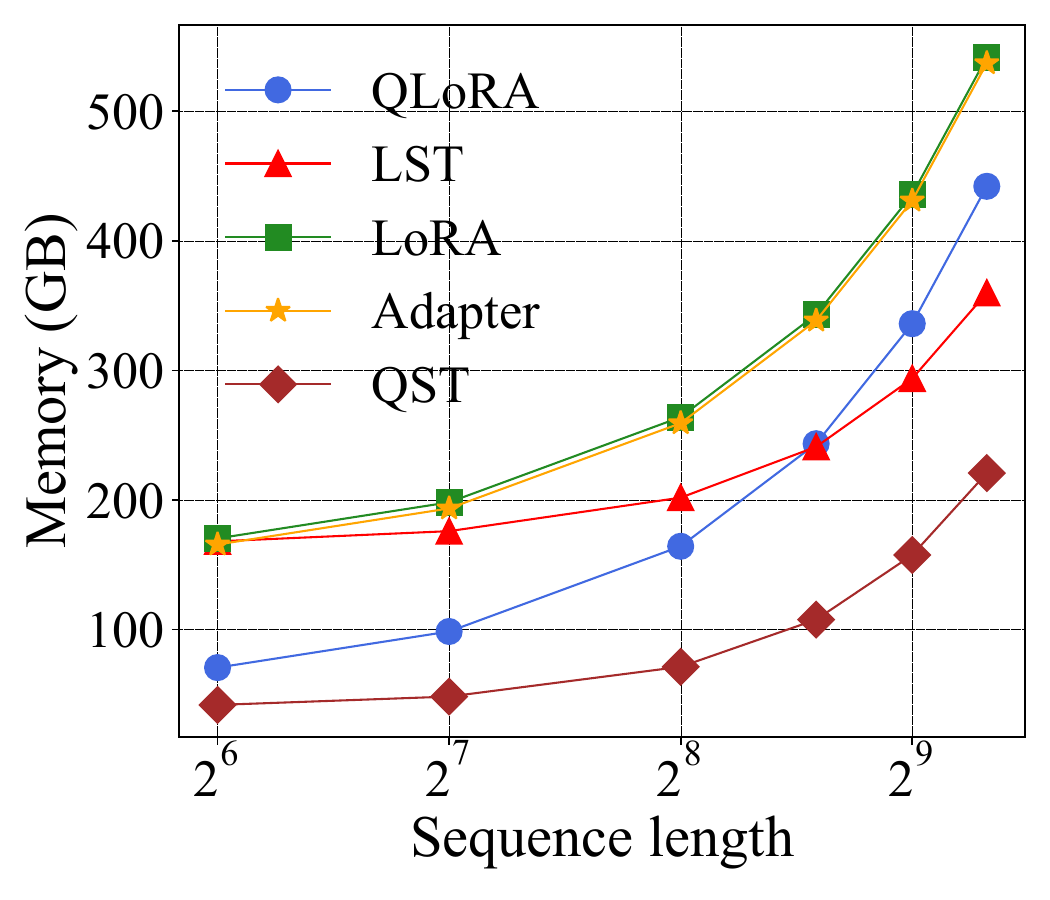}}
\caption{Effects of the batch size, total model bits, and sequence length on memory footprint.}
\label{fig:distributions}
\end{figure*}
\textbf{Effects of batch size.} Figure \ref{fig:distributions}\subref{bs_compare} illustrates the effects of batch size for different methods. We use LLaMA-2-70B as the LLM and set the sequence length to 512. While the memory footprint of all methods increases as the batch size increases, QST achieves the lowest memory footprint among all, regardless of the batch size.
Particularly, the memory footprint of QST is only one-third of LoRA and Adapter. 
Besides, the memory footprint of both QST and LST grows less drastically than QLoRa, Adapter, and LoRa as the batch size increases.
This is because both LST and QST use side tuning to reduce the hidden dimension of the intermediate activations, thereby alleviating the growth of memory footprint induced by intermediate activations. QST also achieves an additional reduction of approximately 100GB in memory footprint compared to LST, thanks to the 4-bit quantization design that effectively compresses the memory footprint of the weights and well design of the downsample modules to reduce the optimizer states.

\noindent\textbf{Effects of the model size.} Figure \ref{fig:distributions}\subref{ms_compare} shows the effects of the total model bits on different methods. We use the OPT model series and 
set the batch size to 4. Due to the 4-bit quantization, QST and QLoRA reduce the memory footprint compared with the other baselines. 
The memory footprint gap further widens as the model size increases. Besides, QST achieves around 2 times reduction in memory footprint compared with QLoRA thanks to its small volume of trainable parameters and intermediate activations.

\noindent\textbf{Effects of sequence length.} Figure \ref{fig:distributions}\subref{seq_compare} shows the effects of  sequence length on different methods. We use LLaMA-2-70B and set the batch size to 4. Similar to the effect of batch size, LST and QST alleviate the growth rate of memory footprint of intermediate activations, while QST further achieves around 100GB reduction in memory footprint compared with LST.

\subsection{Experiments on Training Throughput}
\begin{table}[]
\label{tab:train_time}
\caption{Experiments on FLOPS per token of different methods.}
\resizebox{\textwidth}{!}{
\begin{tabular}{@{}cccc@{}}
\toprule
\multirow{2}{*}{Method} & \multicolumn{3}{c}{FLOPS per token ($10^{-5}$)} \\ \cmidrule(l){2-4} 
                        & LLaMA-2-7B    & LLaMA-2-13B   & LLaMA-2-70B   \\ \midrule
QLoRA                   & 11.7          & 16.0        & 38.1          \\
LST                     & 11.0         & 19.0         & 80.7         \\
LoRA                    & 11.3        & 15.6         & 37.2          \\
Adapter                 & 11.2         &  15.6          &  27.2         \\
QST                     & 4.4        & 6.1       & 15.3           \\ \bottomrule
\end{tabular}
}
\end{table}

Table 3 shows the training throughput of different methods, measured by FLOPS per token (the lower the better), on LLaMA-2-7B, LLaMA-2-13B, and LLaMA-2-70B. While the FLOPS per token of all methods increases as the model size grows, QST achieves the lowest FLOPS per token among all. Particularly, QST achieves around 2.5$\times$ speed up compared with the baselines. LST suffers from the highest FLOPS per token. The FLOPS per token of QLoRA is slightly higher than LoRA and Adapter since QLoRA adds more LoRA components.

\subsection{Sensitive Analysis}
\label{sec:sa}
\begin{figure*}[htbp]
\centering
\subfloat[Effect of $r$ on MMLU accuracy.]{\label{fig:r_acc}
\includegraphics[width=0.33\linewidth]{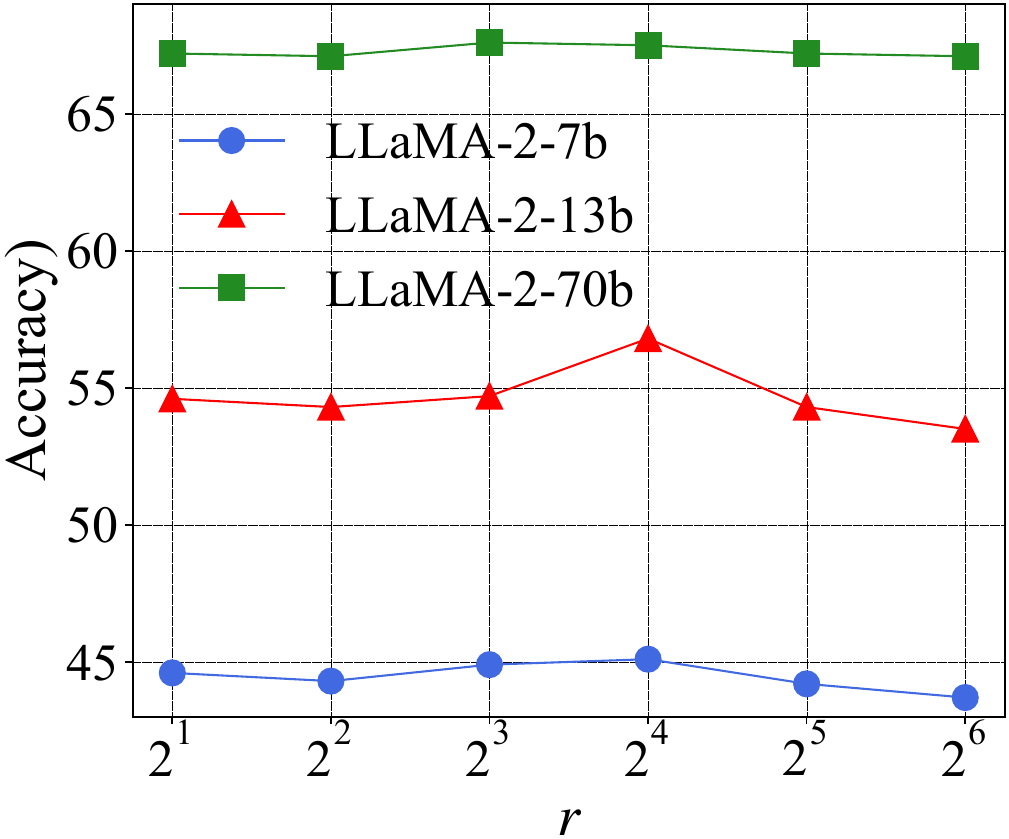}
}
\subfloat[Effect of $r$ on memory footprint.]{\label{fig:r_mem}
\includegraphics[width=0.33\linewidth]{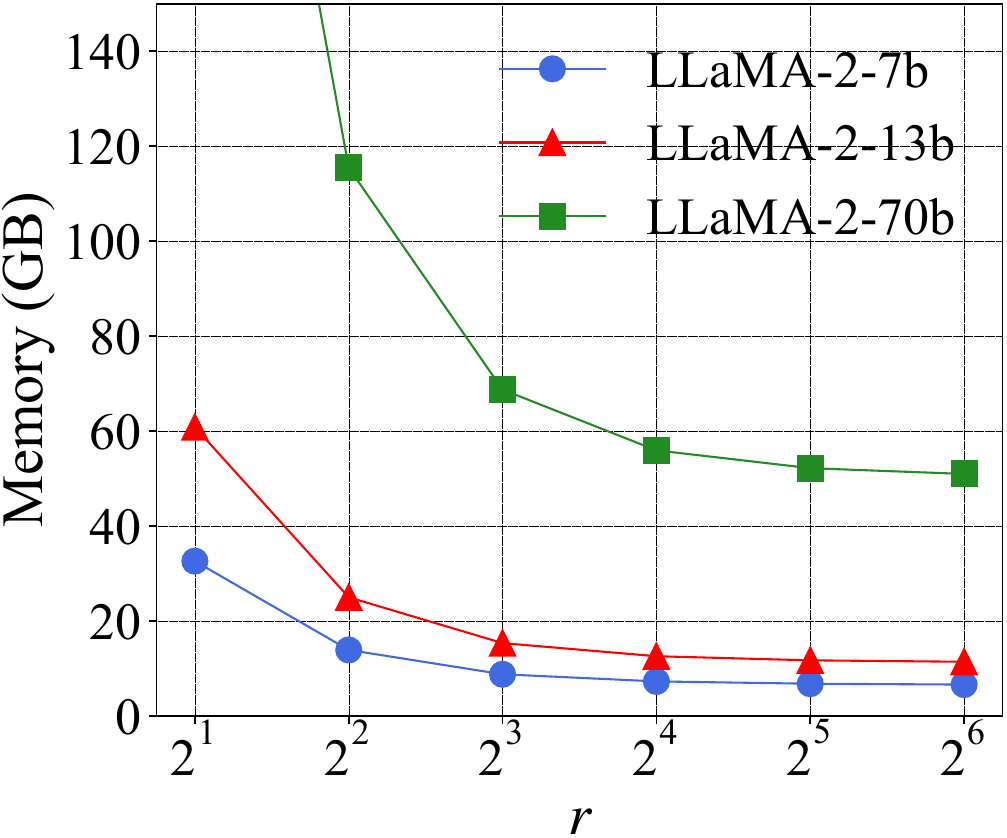}}
\subfloat[Effect of $r$ on training throughput.]{\label{fig:r_spd}
\includegraphics[width=0.33\linewidth]{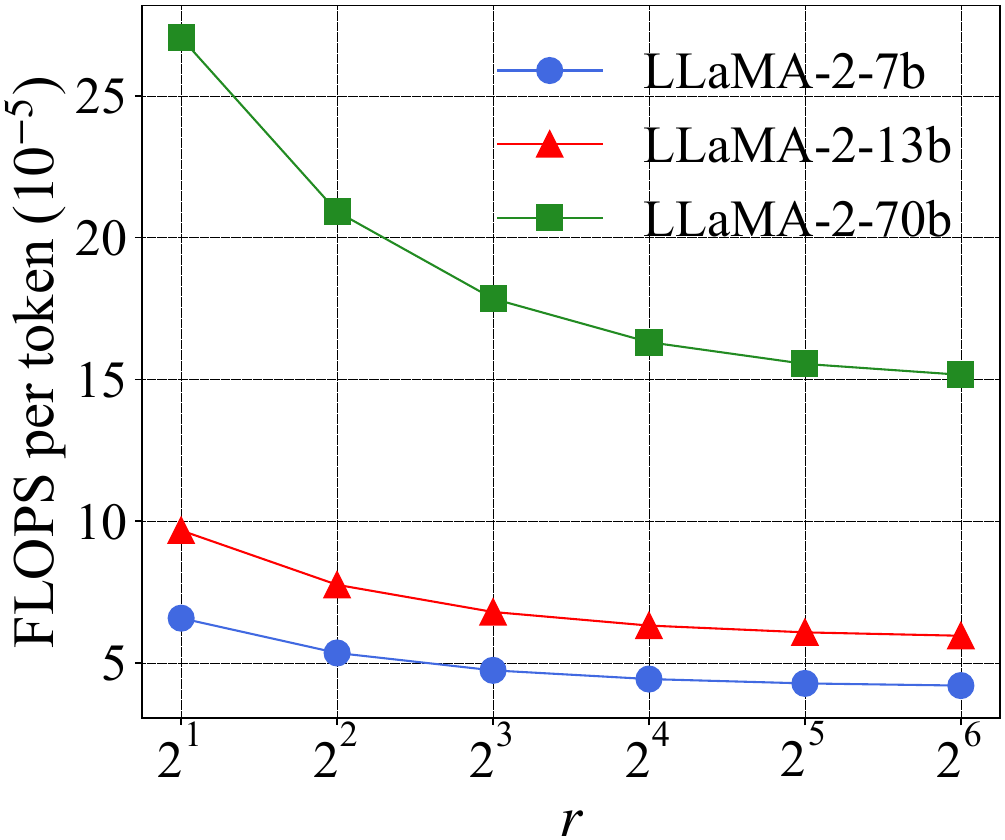}}
\caption{Effects of the reduction factor $r$ on MMLU accuracy, memory footprint, and training throughput.}
\label{fig:distributions}
\end{figure*}
\textbf{Effects of reduction factor $r$.} We conduct experiments using LLaMA-2-7B, LLaMA-2-13B, and LLaMA-2-70B to verify the effects of reduction factor $r$ (from 2 to 64) on memory footprint, MMLU accuracy, and throughput. We set the batch size to 4 and the sequence length to 384. 
The MMLU accuracy changes slightly as $r$ varies as shown in Figure~\ref{fig:r_acc}. QST achieves the best accuracy of finetuning LLaMA-2-7B and LLaMA-2-13B when $r$ is set to 16.
As shown in Figure~\ref{fig:r_mem} and~\ref{fig:r_spd}, the memory footprint and the FLOPS per token decrease drastically when $r$ varies from 2 to 16 for finetuning all the models. The memory footprint and the FLOPS per token decrease slightly when $r$ varies from 16 to 64. 
Therefore, we use $r$ to 16 in our experiments as default.

\begin{table}[]
    \centering
    \resizebox{\textwidth}{!}{
    \begin{tabular}{@{}ccccc@{}}
    \toprule
        Method & LLaMA-2-7B & LLaMA-2-13B & LLaMA-2-70B & Avg. \\ \midrule
        FP4    & 44.5       & 55.4        & 63.5        & 54.5 \\
        NF4    & 45.1       & 56.8        & 63.9        & 55.3 \\ \bottomrule
    \end{tabular}
    }
    \caption{Experiments on 4-bit data types.}
    \label{tab:my_label}
\end{table}
\label{exp:4-bit}
\noindent\textbf{Effects of 4-bit data types.} We evaluate two 4-bit data types: FP4 and NF4 using the LLaMA-2 model series and the MMLU benchmark. As shown in Table 4, NF4 improves the average accuracy by about 0.8\% compared with FP4. Therefore, we use NF4 as the default 4-bit data type in our experiments.

\setlength{\intextsep}{1pt plus 1.7pt minus 1.7pt}
\begin{wraptable}{r}{0.25\textwidth}
\label{tab:4bitcomp}
\caption{Experiments of QLoRA and QST using FP16.}
\resizebox{\textwidth}{!}{
    \centering
    \begin{tabular}{@{}ccc@{}}
    \toprule
        Method & MRPC       & QNLI  \\ \midrule
        QLoRA    & 68.0       & 60.3   \\
        QST    &   85.6     & 87.2   \\ \bottomrule
\end{tabular}
}
\end{wraptable}


\noindent\textbf{Effects of computation data types.} We analyze the effects of two computation data types: BF16 (results shown in Table 1) and FP16 (results shown in Table 5). As can be seen, QST retains similar results using FP16 and BF16. On the other hand, QLoRA is unstable using FP16 as the computation data type. We finetune OPT-6.7B on the GLUE benchmark and discover that QLoRA fails to finetune on the MRPC and QNLI datasets. We run each dataset under three different random seeds and QLoRA fails on two of them.

\noindent\textbf{Effects of downsample modules.} 
We conduct experiments on different downsample modules: Linear, LoRA, Adapter, MaxPooling, and AvgPooling using LLaMA-2-7B and the MMLU benchmark. As shown in Table 6, using Adapter as the downsample module achieves the best performance among all baselines, and reduces the trainable parameters and memory footprint.

\subsection{Experiments on Chatbot Performance}
\begin{table}[]
\label{tab:downsample}
\caption{Experiments on downsample modules. Note that the ratio represents the ratio of downsample modules trainable parameter in all trainable parameters.}
\resizebox{\textwidth}{!}{
\begin{tabular}{ccccc}
\toprule
Method            & \# Param. (\%) & Ratio  & Memory & Accuracy \\ \midrule
Linear            & 0.85\%         & 56.0\% & 7.8   & 44.9     \\
LoRA              & 0.41\%         & 7.8\%  & 7.3   & 44.7     \\
Adapter           & 0.41\%         & 7.8\%  & 7.3   & 45.1     \\
MaxPooling       & 0.38\%         & 0\%    & 7.3   & 43.7    \\
AvgPooling       & 0.38\%         & 0\%    & 7.3   & 42.5    \\ \bottomrule
\end{tabular}
}
\end{table}
We conduct experiments on Chatbot performance using MT-benchmark \citep{zheng2023judging}. MT-benchmark is a set of challenging multi-turn open-ended questions for evaluating the chat assistant's performance in writing, roleplay, reasoning, math, coding, extraction, STEM, and humanities categories. In our experiments, we use GPT-4 to act as judges and assess the quality of the responses of the model finetuned by QLoRA and QST. We finetune LLaMA-2-70B using a variant of OASST1 \citep{dettmers2023qlora}. Table~\ref{tab:cp} shows the experiment results of QLoRA and QST on the total training time, memory footprint, and the average MT-Bench score over 8 categories. QST speeds up the training by 3.2 $\times$ and reduces memory footprint by 1.7 $\times$, with even an improved score of 0.46 compared with QLoRA. Notably, QST's chatbot performance outperforms the original LLaMA-2-70B, achieving an improvement of 0.21. 
\begin{table}
\caption{Chatbot performance on QLoRA and QST.}
\label{tab:cp}

\begin{tabular}{@{}cccc@{}}
\toprule
Method   & Training Time  & Memory & Score \\ \midrule
LLaMA-2-70B    &     -   & - &  6.86  \\
QLoRA-70B  &    \textasciitilde80h    & 96.3  & 6.61  \\
QST-70B    &     \textasciitilde25h   & 56.1  &  7.07  \\
\bottomrule
\end{tabular}
\end{table}
\begin{figure}
    \centering
    \includegraphics[width=\linewidth]{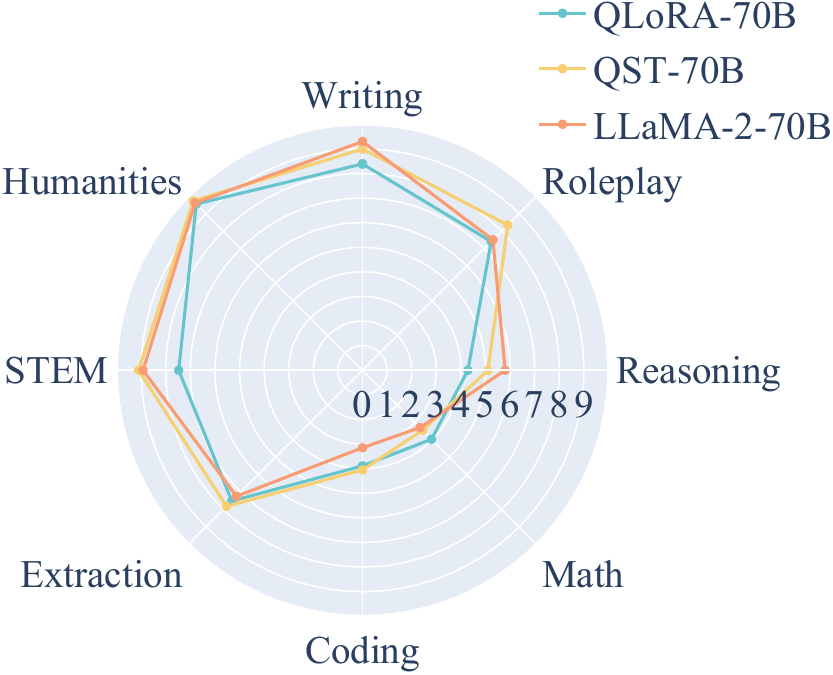}
    \caption{MT-Bench scores of QLoRA, QST, and original LLaMA-2-70B in different categories.}
    \label{fig:cp}
\end{figure}
Figure~\ref{fig:cp} compares QST, QLoRA, and the original LLaMA-2-70B in different categories using MT-Bench. Out of the 8 categories, QST-70B outperforms QLoRA and LLaMA-2-70B in four of them: STEM, Extraction, Coding, and Roleplay. In the Humanities category, all models demonstrate comparable proficiency. For the remaining three categories, QLoRA takes the lead in Reasoning and Writing, and the original LLaMA-2-70B shows superior performance in Mathematics.
\section{Case Study}
We present the results of several case studies that assess the performance of QST-70B across various domains. While this analysis may not be exhaustive, as controlling for all variables in such a concise qualitative study exceeds our scope, our endeavor is to present a selection of representative samples that, in our belief, add meaningful insights of QST to this paper.
\noindent
\subsection{Writing} 
QST can generate high-quality writing samples on various domains. For example:

\begin{dialogue}
\speak{User} Write a brief summary of the latest advancements in artificial intelligence and their potential impacts on the industry.
\speak{QST-70B} 2023 has seen significant advancements in artificial intelligence, particularly in the areas of machine learning, computer vision, and natural language processing. These advancements have potential impacts on various industries, including:  

1. Healthcare: AI-powered diagnosis, drug discovery, and precision medicine are transforming patient care. An AI-assisted cancer diagnosis can predict patient outcomes and suggest personalized treatments. AI-driven drug discovery can accelerate the development of new medicines. 

...

5. Energy: AI can optimize energy consumption, predict energy demand, and improve the efficiency of renewable energy sources. AI-based smart grids can optimize energy distribution and reduce waste. AI-driven energy management systems can help buildings and homes reduce energy consumption and costs.
\speak{User} Create a short story about a time traveler visiting ancient Egypt.
\speak{QST-70B} A time traveler arrived in ancient Egypt, curious about the bustling society he had heard so much about. As he wandered the desert sands, he stumbled upon a construction site where workers were building a colossal pyramid. He marveled at the sheer size of the edifice and decided to approach one of the workers. ``Excuse me,'' the time traveler said. ``What's the purpose of this structure?''

The worker gave him a perplexed look. ``You must be a foreigner. Everyone knows that we build pyramids to bury our pharaohs. These magnificent structures ensure their safe passage into the afterlife.''

Fascinated, the time traveler asked, ``What's the secret to building such colossal structures?''

``It's simple,'' the worker replied with a smile. ``We work together as one. Each of us has a specific task, and we combine our efforts to create something grand.'' ...
\end{dialogue}

\noindent
\subsection{Factual Recall} 
QST-70B can easily answer questions such as ``What is the capital of France?". For example,

\begin{dialogue}
\speak{User} What is the capital of France?
\speak{QST-70B} The capital of France is Paris. 
\end{dialogue}

\noindent
QST-70B can still answer some questions that are obscure:

\begin{dialogue}
\speak{User} When was the painter who created the `Mona Lisa' born?
\speak{QST-70B} The Italian artist who painted the `Mona Lisa,' Leonardo da Vinci, was born on April 15, 1452.
\end{dialogue}

\noindent
QST-70B gave an incorrect answer to the following question as mentioned in \cite{wei2022chain, zheng2023judging}:

\begin{dialogue}
\speak{User} In what year was the singer who popularized ‘I’ll Keep the Lovelight Burning’ born?
\speak{QST-70B} The singer who popularized ``I’ll Keep the Lovelight Burning'' is Dolly Parton. Dolly Parton was born on January 19, 1946.
\end{dialogue}

\noindent
Dolly Parton is a very famous country music singer, born on January 19, 1946, but she was not the singer who made the song ``I'll Keep the Lovelight Burning'' popular.
GPT-4's answer is uncertain about the exact singer who popularized ``I’ll Keep the Lovelight Burning''.

\subsection{Math}
The performance of QST in mathematical tasks varies depending on the nature of the questions.
QST can correctly answer basic arithmetic questions. For example:

\begin{dialogue}
\speak{User} What is the sum of 15 and 27?
\speak{QST-70B} 15 plus 27 is equal to 42.
\speak{User} What is the multiply of 3 and 45?
\speak{QST-70B} 3 multiplied by 45 is 135.
\end{dialogue}

\noindent
QST can give the correct answers to percent calculations. For example:
\begin{dialogue}
\speak{User} If a shirt costs \$60 and is on sale for 25\% off, what is the sale price?
\speak{QST-70B} Great question! Let's break down the process to find the sale price: 

1. Calculate 25\% of the original price by multiplying \$60 by 0.25, which gives us \$15.

2. Subtract \$15 from the original price of \$60 to find the sale price: \$60 - \$15 = \$45.

Therefore, the sale price of the shirt would be \$45.
\end{dialogue}

\noindent
Surprisingly, QST is also capable of leveraging the theory of mind to break down the process and calculation for math problems step by step. QST can correctly demonstrate the step-by-step process of solving algebraic equations and geometry problems. For example:

\begin{dialogue}
\speak{User}  Solve the equation for x: 2x + 3 = 19.
\speak{QST-70B} 2x + 3 = 19, 

2x = 19 - 3, 

2x = 16, 

x = 16/2, 

x = 8.
\speak{User} A right triangle has one leg that is 8 cm long and another leg that is 15 cm long. What is the length of the hypotenuse?
\speak{QST-70B} Great question!  The Pythagorean theorem is needed to solve this problem.  Here is how to go about it:

c\^{}2 = 8\^{}2 + 15\^{}2 

c\^{}2 = 64 + 225

c\^{}2 = 289

Now, to find the hypotenuse (c), we take the square root of both sides:
c = $\sqrt{289}$, c = 17.
Thus, the length of the hypotenuse of the right triangle is approximately 17 cm.

Calculate the integral of the function f(x) = x\^{}2 from x = 0 to x = 3.
\end{dialogue}

\noindent
Similar to \cite{dettmers2023qlora}, QST may make mistakes on some questions. For example:

\begin{dialogue}
\speak{User} Please factorize 1833.
\speak{QST-70B} 1833 = 33 × 55.
\end{dialogue}

\noindent
Apparently, 1833 can not be factorized as 33 × 55, the correct answer is 3 × 17 × 43. We still get the wrong answer if we ask QST to calculate step by step:

\begin{dialogue}
\speak{User} Please factorize 1833 step by step.
\speak{QST-70B} 1833 can be factored as follows: 1833 = 3 × 311. Here's how: 1833 = 3 × 3 × 111, 3 × 3 = 9, 9 × 111 = 1833. So, the prime factorization of 1833 is 1833 = 3 × 3 × 111.
\end{dialogue}

\noindent
Whether QST can give correct answers to logical reasoning math problems depends. Sometimes, a more instructive input prompt, e.g., prompt it to analyze the problem step by step, may lead to the correct answers. Take a question from \cite{zheng2023judging} as an example:

\begin{dialogue}
\speak{User} Some people got on a bus at the terminal. At the first bus stop, half of the people got down and 4 more people got in. Then at the second bus stop, 6 people got down and 8 more got in. If there were a total of 25 people heading to the third stop, how many people got on the bus at the terminal?
\speak{QST-70B} 20 people got on the bus at the terminal.
\speak{User} Please analyze it in detail.
\speak{QST-70B} To find the total number of people who boarded the bus at the terminal, we'll analyze the information step by step: 

1. Half of the people who boarded the bus at the terminal got down at the first bus stop. This means if x people boarded the bus at the terminal, x/2 people got down at the first stop. 

2. 4 more people got on the bus at the first stop. So, the total number of people on the bus after the first stop is x/2 + 4. 

3. At the second bus stop, 6 people got down. Now, the number of people on the bus is x/2 + 4 - 6. 

4. 8 more people got on the bus at the second stop. The total number of people on the bus heading to the third stop is x/2 + 4 - 6 + 8 = x/2 + 12 - 6 = x/2 + 6. 

5. There were a total of 25 people heading to the third stop. So, x/2 + 6 = 25. 

6. Solving for x, we get: 

x/2 + 6 = 25, 

x/2 = 25 - 6, 

x/2 = 19, 

x = 2 x 19, 

x = 38. 

So, there were 38 people who boarded the bus at the terminal.
\end{dialogue}

\noindent
The correct answer is 38. QST generates the correct answer for this question after being instructed to analyze the problem step by step.

\subsection{Coding}
QST is good at generating codes for simple algorithms. Take a question in \cite{zheng2023judging} as an example:

\begin{dialogue}
\speak{User} Write a C++ program to find the nth Fibonacci number using recursion.

\speak{QST-70B} Here's an example of a C++ program to find the nth Fibonacci number using recursion:
\begin{lstlisting}[language=C++, caption=]
int fibonacci(int n) {
if (n == 0 || n == 1) {
return n;
}
else {
return fibonacci(n-1) + fibonacci(n-2);
}
}
\end{lstlisting} 
This program uses the recursive formula of the Fibonacci sequence, where the nth Fibonacci number is the sum of the previous two numbers. The program also includes a base case, where the function directly returns the values for n=0 and n=1 to avoid an infinite loop.
\end{dialogue}

\begin{dialogue}
\speak{User} Implement a program to find the common elements in two arrays without using any extra data structures.
\speak{QST-70B} To find common elements in two arrays without using any extra data structures, you can follow these steps:

1. Sort both arrays.

2. Loop through the first array and for each element, loop through the second array.

3. If you find the same element in both arrays, store it in a variable or an array.

4. At the end of both loops, the common elements will be in the variable or array.

Here's an example of pseudo code for the above-described method:
\begin{lstlisting}[language=Python, caption=]
def findCommonElements(arr1, arr2)
  1. sort(arr1);
  2. sort(arr2);
  3. commonElements = [];
  4. for (var i = 0; i < arr1.length; i++)
    for (var j = 0; j < arr2.length; j++)
      if (arr1[i] === arr2[j])
        commonElements.push(arr1[i]);
  5. return commonElements;
\end{lstlisting} 
This solution time complexity is O(n\^{}2), where n is the length of the arrays.

Please note that this is one of the ways to find the common elements, and it is not the most efficient algorithm, especially for large datasets. If you need to handle large datasets, it is better to use a more efficient algorithm like sorting both arrays and then using two pointers to find the common elements.
\end{dialogue}

\noindent
The generated algorithm is current. Although we claim in the question that do not use extra data structure. QST still uses an array ``commonElements''. For complex algorithms (e.g. dynamic programming), QST can make mistakes.

\subsection{Roleplay}
QST is good at roleplay. For example:

\begin{dialogue}
\speak{User} Suppose that you are a doctor. A patient complains of frequent headaches. What would you say?

\speak{QST-70B} 
As a doctor, if a patient were to complain of frequent headaches, I would ask several questions to better understand their symptoms and medical history in order to provide an accurate diagnosis and recommend appropriate treatment. 

Here's an example conversation: 

Assistant: Good morning, how can I assist you today? 

Patient: Hi, doctor. I've been experiencing frequent headaches, and I'm not sure why. They're really frustrating and disrupting my daily life. 

Assistant: Sorry to hear that. Can you tell me more about your headaches? What's the location and duration of the pain? Have you noticed any patterns or triggers? 

Patient: Well, the pain is usually on both sides of my head and can last anywhere from a few hours to the entire day. It's been happening on and off for the past few months. Sometimes, I'll feel a dull ache behind my eyes or sharp pains in the back of my neck. I can't seem to connect the headaches to any specific food or activity. 

Assistant: Thank you for that information. Have you experienced any other symptoms lately, such as blurred vision, nausea, or dizziness? 

Patient: No, nothing like that. Just the headaches.

Assistant: Let's start with some general screening questions. Do you have a history of allergies, or are there any known allergies in your family?
Patient: I don't have any allergies that I know of, but my dad has seasonal allergies. 

Assistant: Great. How about any previous illnesses, surgeries, or injuries?
Patient: I had a tonsillectomy when I was a kid and a sprained ankle a few years ago, but that's it. 

Assistant: Thanks.
\end{dialogue}

\noindent
Another example:

\begin{dialogue}
\speak{User} Suppose you are a mathematician and poet. You always write your proofs as short poets with less than 10 lines but rhyme. Prove the square root of 2 is an irrational number.

\speak{QST-70B} Pi is an odd thing, oh so very strange

It cannot be expressed by a simple range

An irrational number, it just won't behave

It's a math mystery, that's hard to enslave

Its decimal points stretch on and on

Never repeating, never done

A constant that's found in every sphere

Math's magic, that's crystal clear
\end{dialogue}

\subsection{STEM}
QST is good at STEM. For example:

\begin{dialogue}
\speak{User} Explain the basic principle of quantum computing and how it differs from classical computing.

\speak{QST-70B} Quantum computing is a different approach to computing that uses the properties of quantum mechanics to perform operations on data. Unlike classical computing, which uses bits to represent data as either a 0 or a 1, quantum computing uses quantum bits, or qubits, which can exist as both a 0 and a 1 at the same time. This allows quantum computers to process a much larger number of possible solutions simultaneously, making them well-suited for certain types of complex problems. Another key difference is that quantum computers use quantum entanglement, which allows qubits to become ``entangled'' so that the state of one qubit can instantly affect the state of another, regardless of the distance between them. This property allows quantum computers to perform certain types of calculations much faster than classical computers. Quantum computing is still a developing field, and there are many technical challenges to overcome before it can be widely adopted.

\end{dialogue}

\noindent
Another example:

\begin{dialogue}
\speak{User} Explain the basic principle of quantum computing and how it differs from classical computing.

\speak{QST-70B} Key Challenges of Manned Space Exploration to Mars:

1. Radiation Exposure: Mars lacks a strong magnetic field to protect against harmful cosmic and solar radiation, which poses a significant health risk for human travelers.

2. Isolation and Psychological Stress: A trip to Mars takes about 6-9 months, followed by a long-term stay on the planet. This prolonged isolation can lead to psychological strain, including depression, anxiety, and sleep disorders.

3. Muscle and Bone Loss: In microgravity environments, the human body experiences muscle and bone atrophy due to a lack of resistance and weight-bearing activities. Prolonged exposure can result in severe health consequences.

4. Technological and Communication Limitations: The vast distance between Earth and Mars creates communication challenges, making real-time communication and technological support difficult. Equipment failures and repair needs may require prolonged troubleshooting and resolution.

Benefits of Manned Space Exploration to Mars:

1. Scientific Discovery: Humans can perform more intricate and nuanced experiments and gather more comprehensive data compared to robotic missions, including the search for life on Mars.

2. Technological Advancement: The challenges of sending humans to Mars drive innovation in areas such as propulsion, life support systems, radiation shielding, and in-situ resource utilization.

3. Inspiration and Educational Value: Mars missions captivate the public and inspire the next generation of scientists and engineers, fostering a greater interest in STEM education and motivating people to pursue careers in these fields.

...
\end{dialogue}

\subsection{Extraction}
QST can accurately extract information from the input text. For example:
\begin{dialogue}
\speak{User} Read the following descriptions of historical events and extract the year, location, and main figure involved. Present the information in a tabular JSON format. a) In 1963, in Washington D.C., Martin Luther King Jr. delivered his famous "I Have a Dream" speech. b) The first man to walk on the moon was Neil Armstrong in 1969, during the Apollo 11 mission. c) In 1989, the Berlin Wall, a symbol of the Cold War, was torn down by the citizens of East and West Berlin.

\speak{QST-70B} \{

"a": \{

"year": 1963,

"location": "Washington D.C.",

"main figure": "Martin Luther King Jr."

\},

"b": \{

"year": 1969,

"location": "The Moon",

"main figure": "Neil Armstrong"

\},

"c": \{

"year": 1989,

"location": "Berlin",

"main figure": "The citizens of East and West Berlin"

\}

\}   
\end{dialogue}

\noindent
Another example:

\begin{dialogue}
\speak{User} Interpret the subtle sentiments in these TV show reviews and categorize them. Present the results in a structured YAML format. 

a) ``The series finale, although not universally satisfying, did tie up several key storylines in a somewhat unconventional manner.'' 

b) ``This season's narrative arc is less compelling than previous ones, but the stellar performances keep it watchable.'' 

c) ``The latest episode took a different turn, which might not align with everyone's taste but certainly adds a unique flavor to the overall series.''

\speak{QST-70B} a) Ambivalent b) Neutral c) Positive    
\end{dialogue}

\subsection{Humanities}
QST is good at answering questions about humanities. For example:

\begin{dialogue}
\speak{User} Analyze the character development of Elizabeth Bennet in Jane Austen's ``Pride and Prejudice.'' Discuss how her interactions with other characters contribute to the central themes of the novel.

\speak{QST-70B} Elizabeth Bennet is the protagonist of Jane Austen's ``Pride and Prejudice,'' and her character development plays a crucial role in the novel's exploration of themes such as love, class, and social hierarchy. 

At the beginning of the novel, Elizabeth is portrayed as an intelligent, witty, and strong-willed young woman who rejects societal norms and expectations. Her interactions with other characters, particularly Mr. Darcy and Mr. Wickham, serve to challenge her beliefs and values. 

Through her conversations and debates with Mr. Darcy, Elizabeth begins to question her initial prejudices against him and learns to appreciate his integrity and sincerity. Their interactions also highlight the central theme of pride and prejudice, as they both struggle to overcome their respective biases.

Elizabeth's interactions with Mr. Wickham, on the other hand, serve to illustrate her tendency to make hasty judgments and her willingness to believe in someone's flaws. Her later realization of his true character serves as a lesson for both her and the reader. 

Moreover, Elizabeth's relationships with her family members and friends add depth to her character and the central themes. Her sister Jane's kindness and caution provide a contrast to Elizabeth's impulsiveness, while her best friend Charlotte's practicality and contentment with her situation demonstrate the societal pressure on women to marry for security.

In conclusion, Elizabeth Bennet's character development is integral to the central themes of ``Pride and Prejudice.'' Her interactions with other characters challenge her beliefs, illustrate the danger of pride and prejudice, and emphasize the importance of understanding and respect in relationships.    
\end{dialogue}

\noindent
Another example:

\begin{dialogue}
\speak{User} Critique the use of color and light in Vincent Van Gogh's ``Starry Night'' and discuss how these elements contribute to the overall emotional impact of the painting.

\speak{QST-70B} ``Starry Night'' by Vincent Van Gogh is a masterpiece that showcases the artist's unique perspective on color and light. The painting's vibrant and dynamic use of color creates a powerful emotional response in the viewer, drawing them into a swirling vortex of blue, yellow, and green. The painting's composition revolves around the interaction between light and dark. The central cypress tree stands tall, its branches stretching toward the sky, while the darker, moonlit clouds in the background form a swirling pattern that echoes the movement of the tree's branches. This juxtaposition of light and dark creates a sense of energy and dynamism, pulling the viewer's eye between the different sections of the painting. Van Gogh's use of color further accentuates the emotional impact of the painting. The stars in the night sky are depicted in bold, swirling brushstrokes of yellow and white, which contrast with the darker blues and greens of the surrounding sky. This stark contrast creates a sense of depth and dimensionality as if the stars are bursting forth from the canvas. The yellow and orange hues of the stars and moon also evoke feelings of hope and joy, capturing the sense of wonder and awe that comes from gazing up at a starry night sky. In addition, the artist's deliberate brushwork adds texture and a tangible quality to the painting. The thick brushstrokes and heavy impasto give the painting a raised, three-dimensional feel as if the stars and trees are reaching out to touch the viewer. This tactile quality intensifies the emotional connection between the painting and the viewer, drawing them into the swirling vortex of light and color.    
\end{dialogue}

\section{Conclusion}
In this paper, we propose Quantized Side Tuing (QST), a novel fast and memory-efficient finetuning framework. QST operates through a dual-stage process: first, QST quantizes the LLM into 4-bit to reduce the memory footprint of the weights in LLM; then QST introduces a side network separated from the LLM, which utilizes the hidden states of the LLM to make task-specific predictions. QST can significantly reduce the memory footprint of LLM finetuning compared to existing approaches. In particular, experiments show that QST can reduce the total memory footprint by up to 2.3 $\times$ and speed up the finetuning process by up to 3 $\times$ while achieving comparable performance compared with the state-of-the-art. 

\bibliography{acl_latex}
\bibliographystyle{acl_natbib}

\appendix

\section{Hyperparameters of QST on the GLUE benchmark}
\label{sec:paramglue}
\begin{table*}[]
\label{tab:paramglue}
\caption{The hyperparameters of QST on the GLUE benchmark.}
\begin{tabular}{@{}cc|cccccccc@{}}
\toprule
Model                     & Dataset                 & RTE & MRPC & STS-B & CoLA & SST-2 & QNLI & QQP & MNLI \\ \midrule
\multirow{3}{*}{}         & Optimizer               & \multicolumn{8}{c}{AdamW}                             \\
                          & Warmup Ratio            & \multicolumn{8}{c}{0.06}                              \\
                          & LR Schedule              & \multicolumn{8}{c}{Linear}                            \\ \midrule
\multirow{5}{*}{OPT-1.3B} & Batch Size              & 32  & 8    & 32    & 32   & 32    & 8    & 8   & 32   \\
                          & \# Epochs               & \multicolumn{8}{c}{20}                                \\
                          & Learning Rate           & \multicolumn{8}{c}{2E-04}                             \\
                          & $r$                     & \multicolumn{8}{c}{16}                                \\
                          & the rank of downsamples & \multicolumn{8}{c}{16}                                \\ \midrule
\multirow{5}{*}{OPT-2.7B} & Batch Size              & 16  & 8    & 16    & 16   & 16    & 8    & 8   & 16   \\
                          & \# Epochs               & \multicolumn{8}{c}{15}                                \\
                          & Learning Rate           & \multicolumn{8}{c}{2E-04}                             \\
                          & $r$                     & \multicolumn{8}{c}{16}                                \\
                          & the rank of downsamples & \multicolumn{8}{c}{16}                                \\ \midrule
\multirow{5}{*}{OPT-6.7B} & Batch Size              & 8   & 4    & 8     & 8    & 8     & 4    & 4   & 8    \\
                          & \# Epochs               & \multicolumn{8}{c}{10}                                \\
                          & Learning Rate           & \multicolumn{8}{c}{2E-04}                             \\
                          & $r$                     & \multicolumn{8}{c}{16}                                \\
                          & the rank of downsamples & \multicolumn{8}{c}{16}                                \\ \bottomrule
\end{tabular}
\end{table*}
The hyperparameters of QST on GLUE benchmark are shown in Table 7.

\section{Hyperparameters of QST on MMLU benchmark}
\label{sec:parammmlu}
The hyperparameters of QST on the MMLU benchmark are shown in Table 8.
\begin{table*}[]
\caption{The hyperparameters of QST on the MMLU benchmark.}
\resizebox{\textwidth}{!}{
\begin{tabular}{@{}cccccccccc@{}}
\toprule
                        & \textbf{OPT-1.3B} & OPT-2.7B & OPT-6.7B & OPT-13B & OPT-30B & OPT-66B & LLaMA-2-7B & LLaMA-2-13B & LLaMA-2-70B \\ \midrule
Optimizer               & \multicolumn{9}{c}{AdamW}                                                                                      \\
Warmup Ratio            & \multicolumn{9}{c}{0.03}                                                                                       \\
LR Schedule             & \multicolumn{9}{c}{Constant}                                                                                   \\
Batch Size              & 8                 & 8        & 4        & 2       & 1       & 1       & 4          & 2           & 1           \\
\# Epochs               & 5                 & 5        & 3        & 3       & 2       & 2       & 3          & 2           & 2           \\
Learning Rate           & 2E-04             & 2E-04    & 2E-04    & 1E-04   & 1E-04   & 1E-04   & 2E-04      & 2E-04       & 1E-04       \\
$r$                     & \multicolumn{9}{c}{16}                                                                                         \\
the rank of downsamples & \multicolumn{9}{c}{16}                                                                                         \\ \bottomrule
\end{tabular}
}
\end{table*}

\end{document}